\documentclass[10pt,twocolumn,letterpaper]{article}

\usepackage[pagenumbers]{wacv} % To force page numbers, e.g. for an arXiv version

\usepackage{graphicx}
\usepackage{amsmath}
\usepackage{amssymb}
\usepackage{booktabs}
\usepackage{nth}
\usepackage{enumitem}
\usepackage{graphicx}
\usepackage{multirow}
\usepackage[dvipsnames]{xcolor}

\usepackage[pagebackref,breaklinks,colorlinks]{hyperref}

\usepackage[capitalize]{cleveref}
\crefname{section}{Sec.}{Secs.}
\Crefname{section}{Section}{Sections}
\Crefname{table}{Table}{Tables}
\crefname{table}{Tab.}{Tabs.}

\def\wacvPaperID{3} % *** Enter the WACV Paper ID here
\def\confName{WACV}
\def\confYear{2025}

\begin{document}

%%%%%%%%% TITLE - PLEASE UPDATE
\title{Zero-shot hazard identification in Autonomous Driving: \\ {A Case Study on the COOOL Benchmark}}

\author{Lukas Picek$^{1,2,3}$, Vojtěch Čermák$^{1,4}$, and Marek Hanzl$^{1,2}$\\
$^1$ PiVa AI, $^2$University of West Bohemia, $^3$INRIA, $^4$Czech Technical University in Prague \\
{\tt\small lukaspicek@gmail.com / lpicek@inria.cz / picekl@kky.zcu.cz} \\
}
\maketitle

%%%%%%%%% ABSTRACT
\begin{abstract}
This paper presents our submission to the COOOL competition, a novel benchmark for detecting and classifying out-of-label hazards in autonomous driving. 
Our approach integrates diverse methods across three core tasks: (i) driver reaction detection, (ii) hazard object identification, and (iii) hazard captioning. 
We propose kernel-based change point detection on bounding boxes and optical flow dynamics for driver reaction detection to analyze motion patterns. 
For hazard identification, we combined a naive proximity-based strategy with object classification using a pre-trained ViT model. 
At last, for hazard captioning, we used the MOLMO vision-language model with tailored prompts to generate precise and context-aware descriptions of rare and low-resolution hazards. 
The proposed pipeline outperformed the baseline methods by a large margin, reducing the relative error by 33\%, and scored 2nd on the final leaderboard consisting of 32 teams.
\end{abstract}

%%%%%%%%% BODY TEXT
\section{Introduction}
\label{sec:intro}

Autonomous driving has the potential to transform transportation by enhancing safety, reducing traffic accidents, and improving mobility \cite{DiLillo2024,razi2023deep}. Advances in artificial intelligence, machine learning, and sensor systems have enabled autonomous vehicles to navigate complex road scenarios effectively \cite{hu2023planning,chib2023recent}. However, reliably detecting, interpreting, and responding to unforeseen hazards remains a critical barrier since gathering all possible scenarios needed for fully supervised training is impossible \cite{zhao2024autonomous,alshami2024coool}.

Existing autonomous driving systems excel at recognizing predefined objects and events, such as cars, pedestrians, etc., as those are the labels in widely used large-scale datasets (e.g., KITTI\cite{garcia2017review} and nuScenes\cite{caesar2020nuscenes}). However, these benchmarks primarily address known hazards, leaving systems vulnerable to Out-Of-Distribution (OOD) objects and scenarios that require immediate and appropriate responses. The inability to handle these novel challenges effectively can lead to catastrophic consequences, highlighting the need to develop robust and adaptive solutions.

\begin{figure}[t]
    \centering
    \vspace{0.5cm}
    \includegraphics[width=\linewidth]{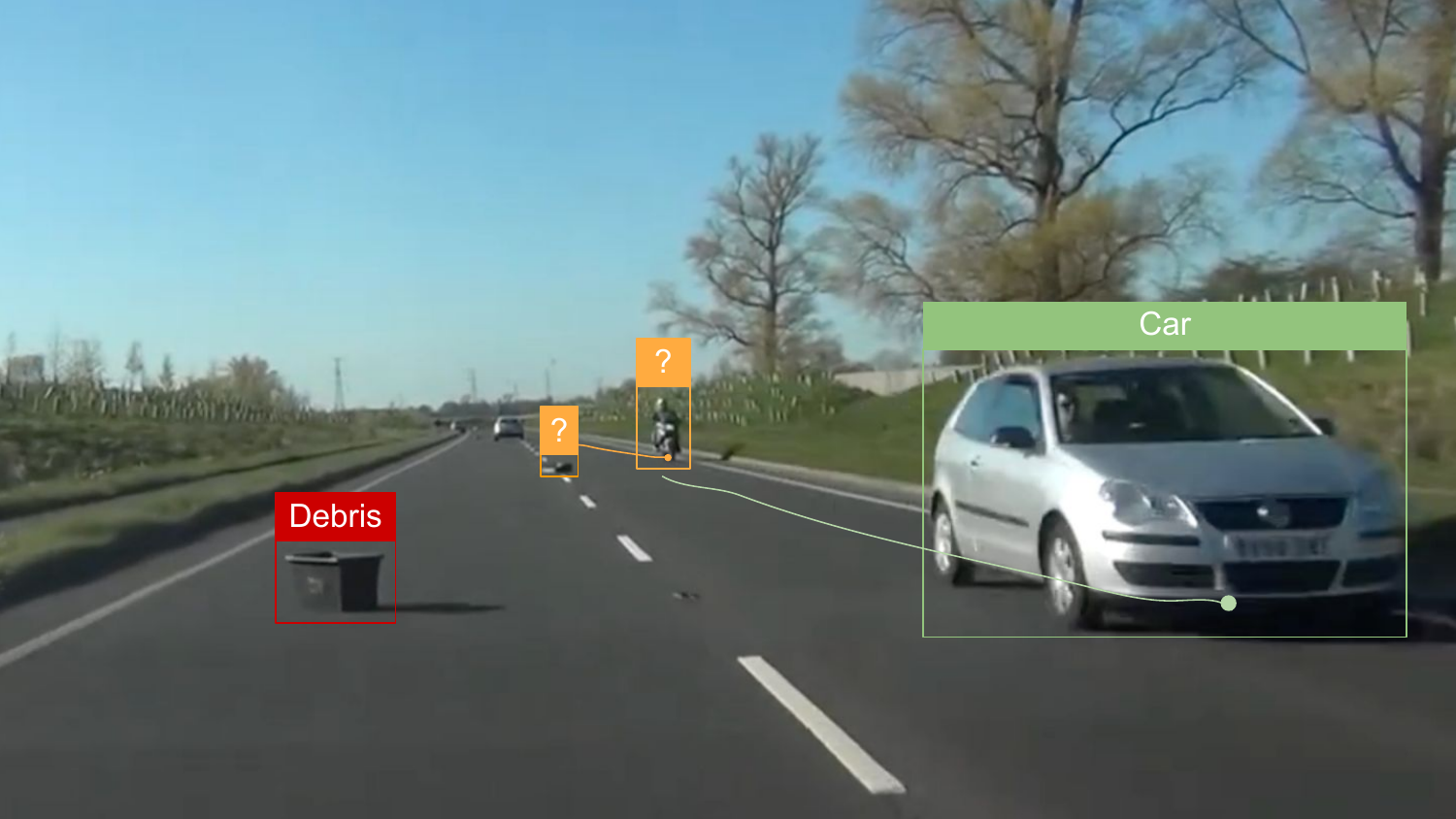}
    \caption{COOOL benchmark focus on zero-shot identification of driver reaction, hazardous objects "recognition," and hazard captioning. A simplified result of our approach is displayed on the selected frame from one of the testing videos. Colors depict the hazard state of each object. Classification is above the object.}
    \label{fig:abstract}
\end{figure}
 
To address this limitation, the COOOL (Challenge Of Out-Of-Label) competition\cite{alshami2024coool} introduces a novel benchmark explicitly designed to evaluate systems' ability to detect and recognize out-of-label hazards. The competition combines high- and low-resolution, annotated dashcam videos focusing on diverse and rare hazards, including exotic animals, debris, and low-resolution obstacles, which are often overlooked in conventional datasets.
Moreover, the competition emphasizes the importance of comprehensive evaluation metrics that reflect real-world scenarios (see Figure \ref{fig:abstract} for reference). Tasks such as detecting the moment of driver reaction, identifying hazardous objects, and accurately classifying hazards are integrated to ensure a holistic system performance assessment. 

This report outlines our participation in the COOOL competition, detailing our approach, insights into the dataset and metrics, and contributions to addressing the challenges posed by novel road scenarios.

\section{Related Work}

The progress in autonomous driving has been shaped the availability of comprehensive datasets and benchmarks, which provide multimodal data to support critical tasks such as object detection, semantic segmentation, and trajectory prediction \cite{garcia2017review,caesar2020nuscenes,sun2020scalability}.
Datasets, such as KITTI, set the stage for benchmarking 2D and 3D perception tasks, offering stereo imagery and LiDAR point clouds to allow research into urban driving scenarios \cite{garcia2017review}. Subsequent efforts like Waymo Open Dataset \cite{sun2020scalability}, nuScenes \cite{caesar2020nuscenes}, and Lyft Level\,5\,\cite{li2023large} expanded the scope by including high-resolution sensors, larger geographical coverage, and a greater variety of driving conditions, such as weather, lighting, and road types. These datasets have played a key role in advancing the state of the art in autonomous navigation, particularly for identifying and managing hazards in well-defined and structured environments. However, their reliance on predefined categories and static taxonomies has revealed inherent limitations when dealing with the unpredictability of real-world road scenarios, where unknown or out-of-label hazards, e.g., fallen debris, exotic animals, or unanticipated pedestrian behaviors, pose a challenge \cite{alshami2024coool, henriksson2023evaluation}.

To mitigate these gaps, \textbf{Open-Set Recognition (OSR)} addresses the challenge of recognizing instances from unknown classes not seen during training \cite{vaze2022openset}. Foundational methods like OpenMax \cite{bendale2016towards} and Extreme Value Machines (EVM) \cite{rudd2017extreme} have inspired advanced techniques such as PostMax, which uses deep feature magnitudes and Generalized Pareto Distributions for probabilistic outputs \cite{cruz2025operational}. Although these methods improve hazard recognition, selecting optimal thresholds for deployment and scaling evaluation remain critical challenges.

Complementary to Open Set Recognition (OSR), \textbf{Out-of-Distribution} (OOD) Detection distinguishes in-distribution samples from those outside the known distribution. Unlike OSR, which assumes unknown data is similar to the training distribution, OOD detection rejects inputs from entirely different distributions.
A common approach involves analyzing classification model outputs to assess whether an image is out of distribution\cite{hendrycks2017a,vaze2022openset}. Enhanced techniques like TRust Your GENerator (TRYGEN) have introduced generative models to detect Out-of-Model Scope hazards by combining original and synthesized features \cite{divivs2024trust}.

In the context of autonomous driving, OOD detection is vital for identifying rare or novel obstacles, such as exotic animals or road debris, that are unlikely to appear in training datasets. Recent advances in OOD \cite{vojivr2025pixood}, using both Vision Transformers and Convolutional Neural Networks, have demonstrated improved large-scale OOD detection capabilities. Yet, like OSR, OOD detection faces the challenge of developing datasets and benchmarks that fully capture the complexity and diversity of real-world hazards. \\

OOD detection and OSR together enable robust autonomous systems by ensuring safe handling of outliers and providing meaningful decision-making in presence of previously unseen objects. The COOOL benchmark serves as a unifying platform, integrating these methodologies to address novel hazards effectively. By including diverse, low-resolution challenges, COOOL pushes the boundaries of existing OOD and OSR techniques, driving innovation and improving their applicability to real-world autonomous systems.

\section{COOOL Challenge}

The COOOL competition \cite{alshami2024coool} (Challenge Of Out-Of-Label) aims to advance autonomous driving research by tackling the challenge of detecting and responding to out-of-label hazards that traditional datasets and methods often overlook. By emphasizing novel scenarios, COOOL bridges the gap between theoretical innovation and real-world application, enabling progress in anomaly detection, open-set recognition, and hazard prediction. Structured around three distinct tasks -- (i) \textbf{Driver Reaction Detection}, (ii) \textbf{Hazard Recognition}, and (iii) \textbf{Hazard Captioning/Classification}  -- the competition encourages the development of algorithms capable of addressing the complexities of dynamic road environments. Its diverse dataset, designed to encompass a wide range of hazards, and rigorous evaluation framework provide the foundation for state-of-the-art advancements that enhance robustness, safety, and adaptability in autonomous systems.

The first focus, \textbf{Driver Reaction Detection}, evaluates a system’s ability to pinpoint the exact moment a driver reacts to a hazard, emphasizing the importance of analyzing subtle visual and temporal cues like changes in vehicle trajectory or driver gaze. Accurate reaction detection is crucial for improving safety and understanding human responses to unforeseen events.

The second focus, \textbf{Hazard Recognition}, assesses the identification and localization of hazardous objects in a scene. This task addresses the complexity of real-world environments, where hazards may include common obstacles as well as rare and unconventional objects like debris or wildlife. Robust algorithms must handle low-resolution, occluded, or dynamic objects, ensuring adaptability to diverse road scenarios.

The third focus, \textbf{Hazard Captioning/Classification}, evaluates systems' ability to classify hazards and assign meaningful labels, whether for known or novel objects. By advancing open-set recognition and novelty detection, this task supports the development of systems capable of understanding complex scenes and handling out-of-distribution samples, ultimately enabling informed decision-making in autonomous vehicles.

\begin{figure*}[t]
    \centering
    \includegraphics[width=0.24\linewidth]{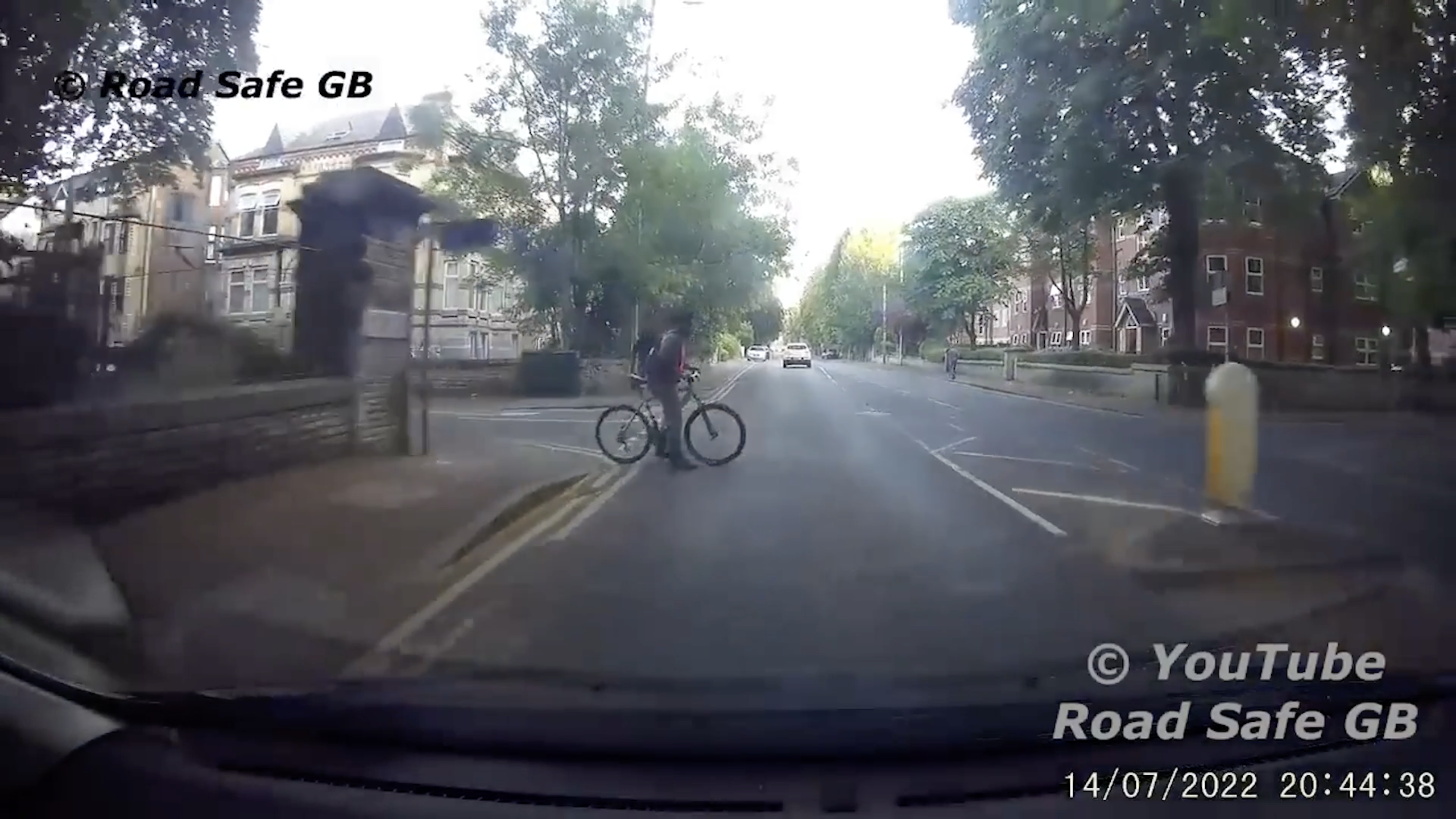}
    \includegraphics[width=0.24\linewidth]{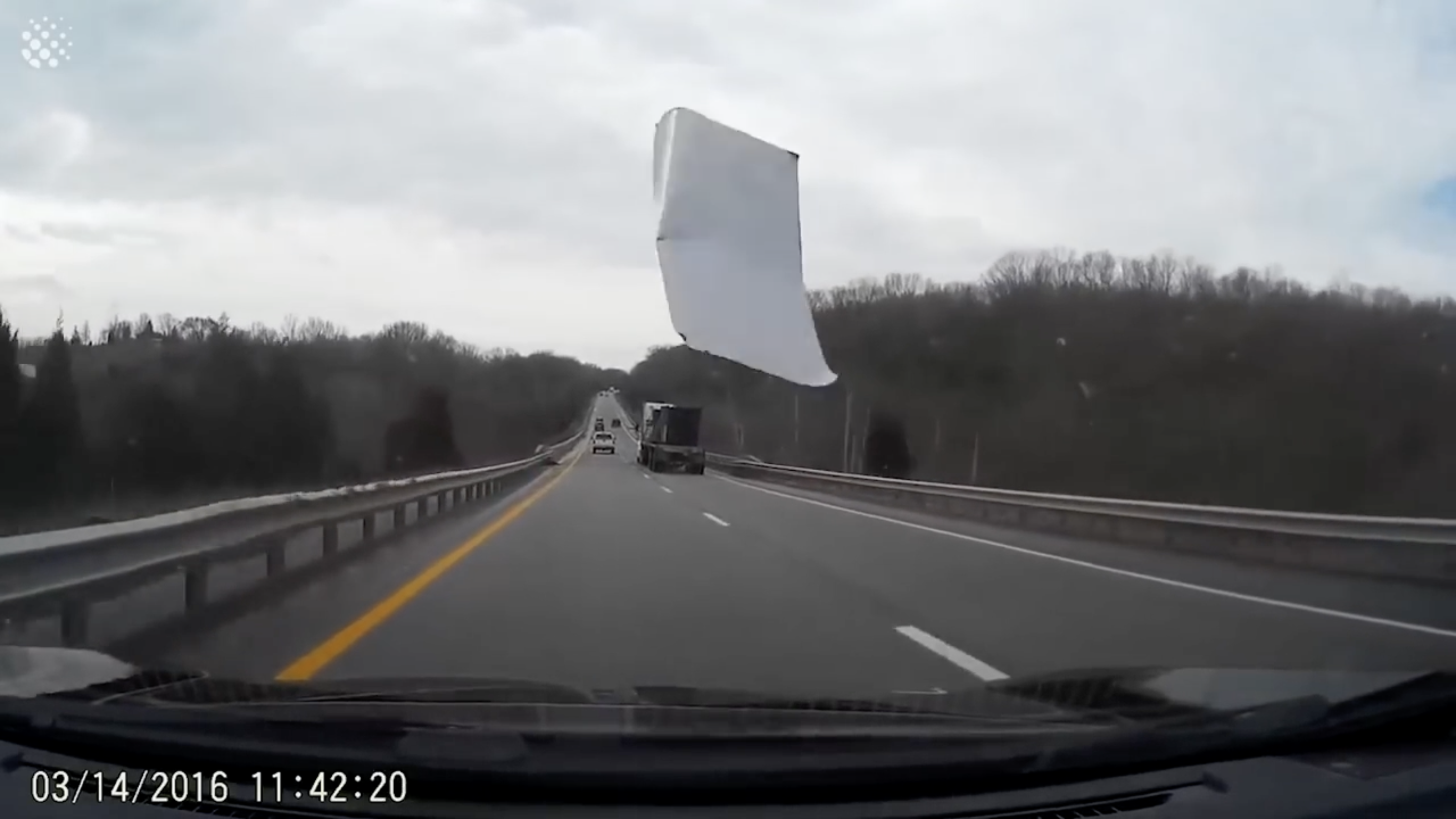}
    \includegraphics[width=0.24\linewidth]{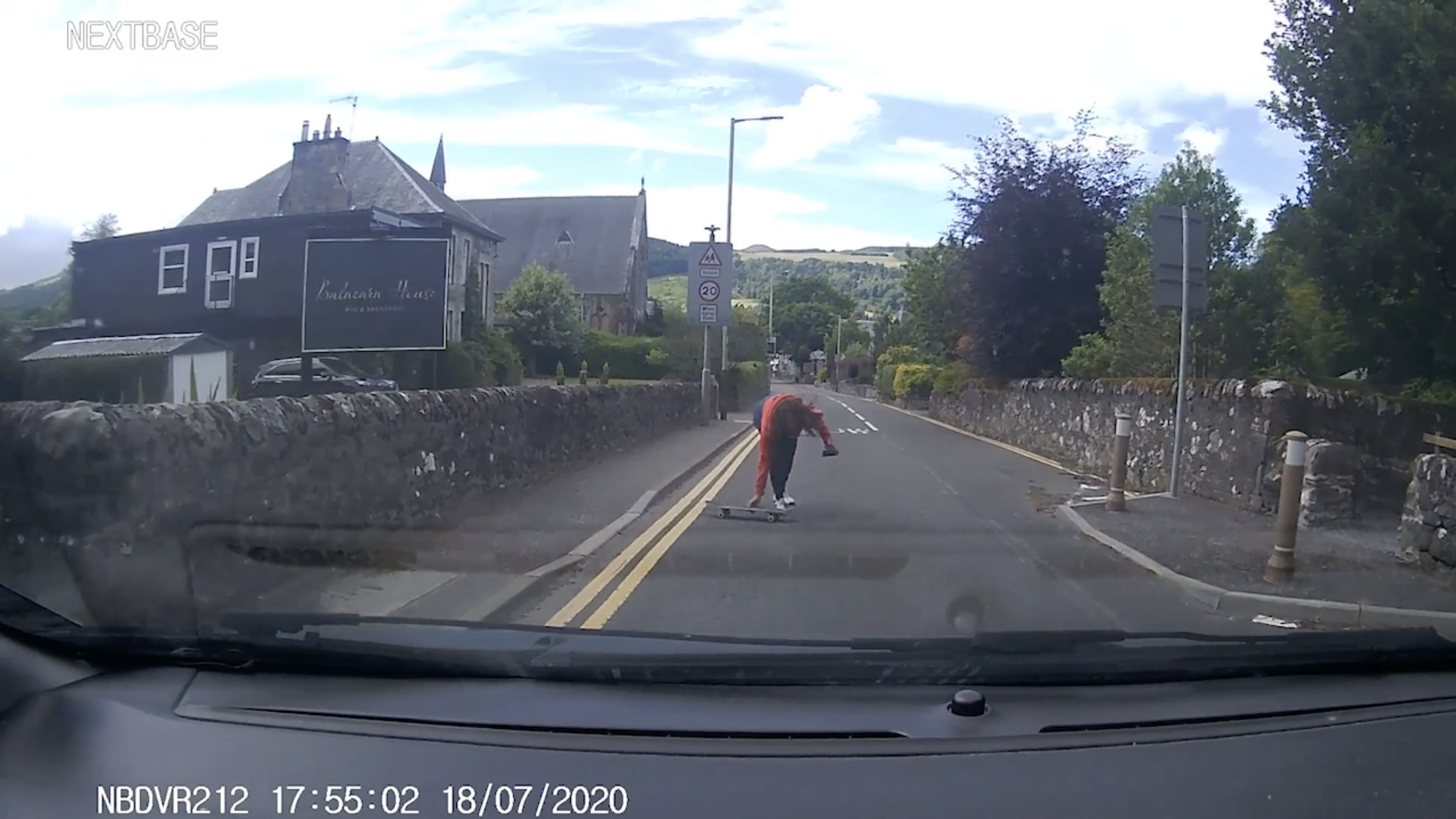}
    \includegraphics[width=0.24\linewidth]{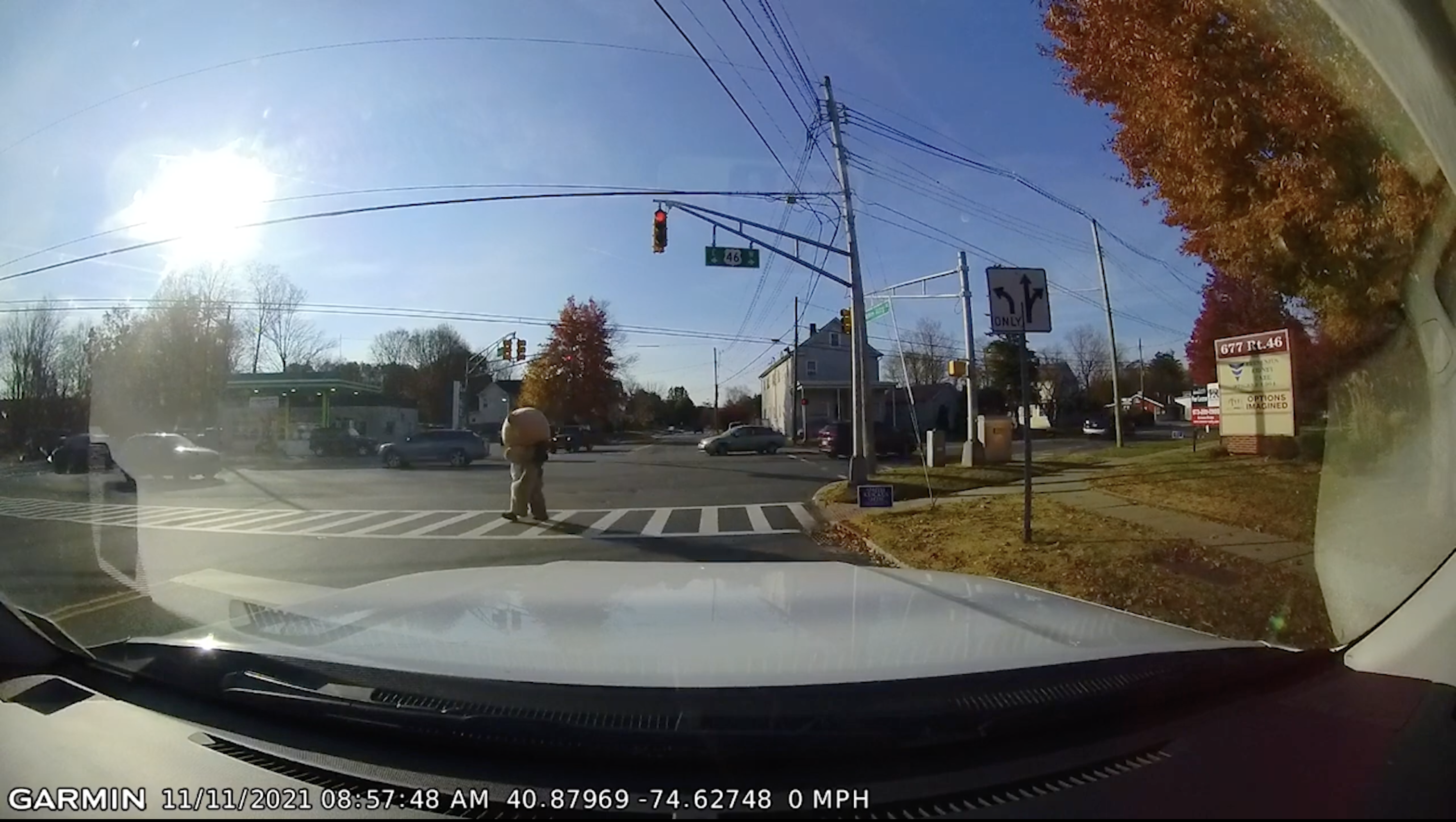} \\
    \vspace{1px}
    \includegraphics[width=0.24\linewidth]{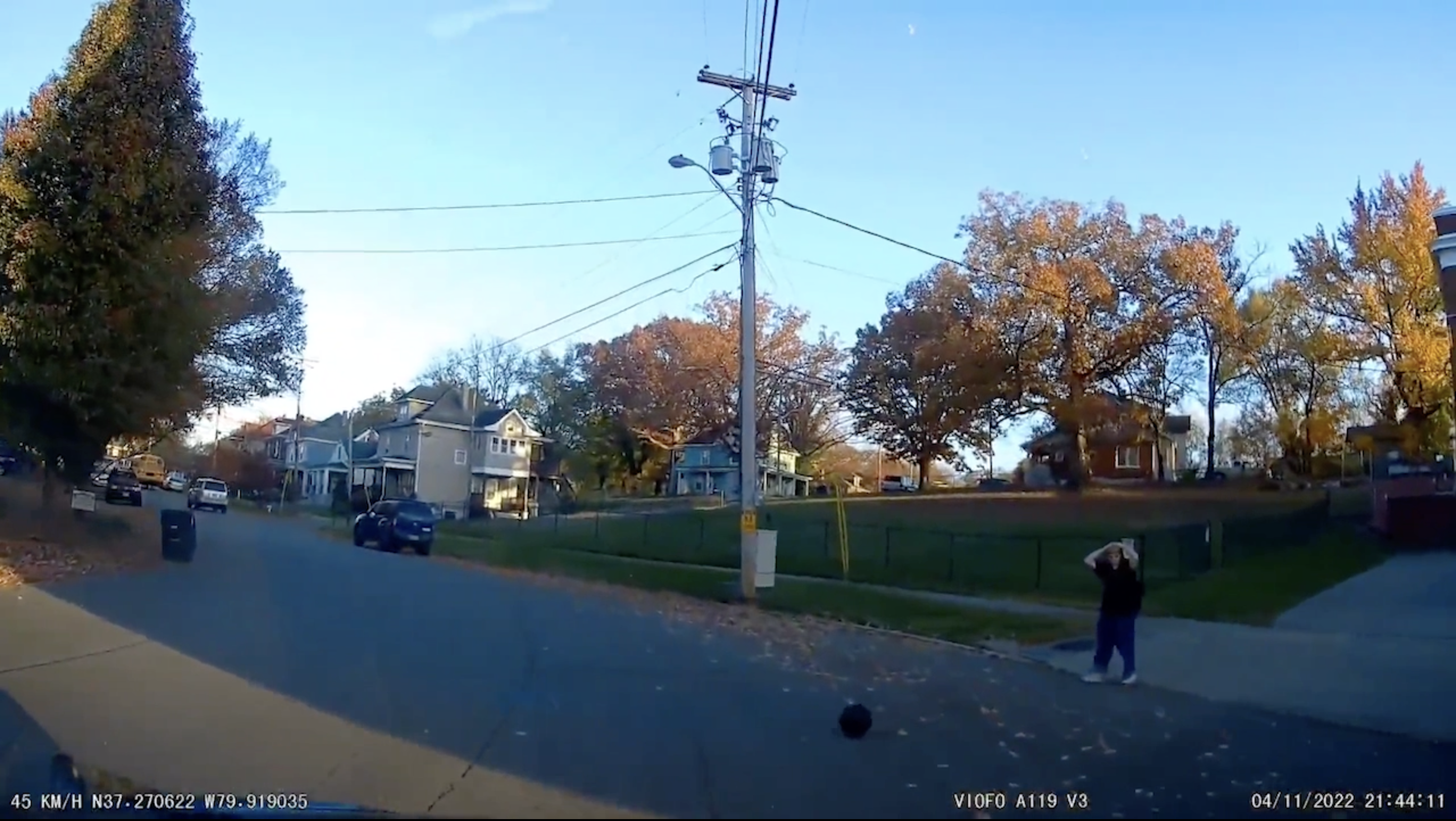} 
    \includegraphics[width=0.24\linewidth]{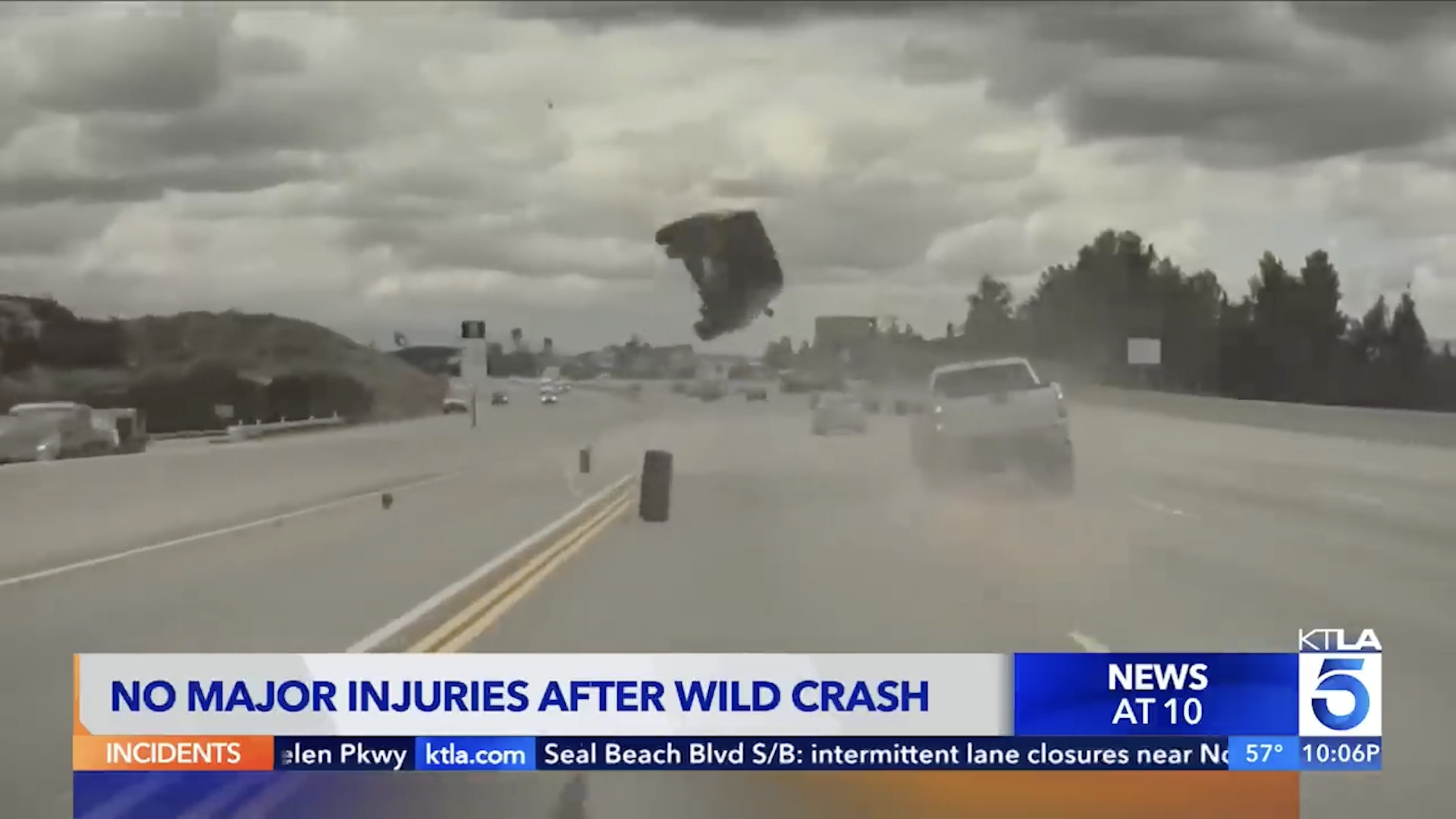} 
    \includegraphics[width=0.24\linewidth]{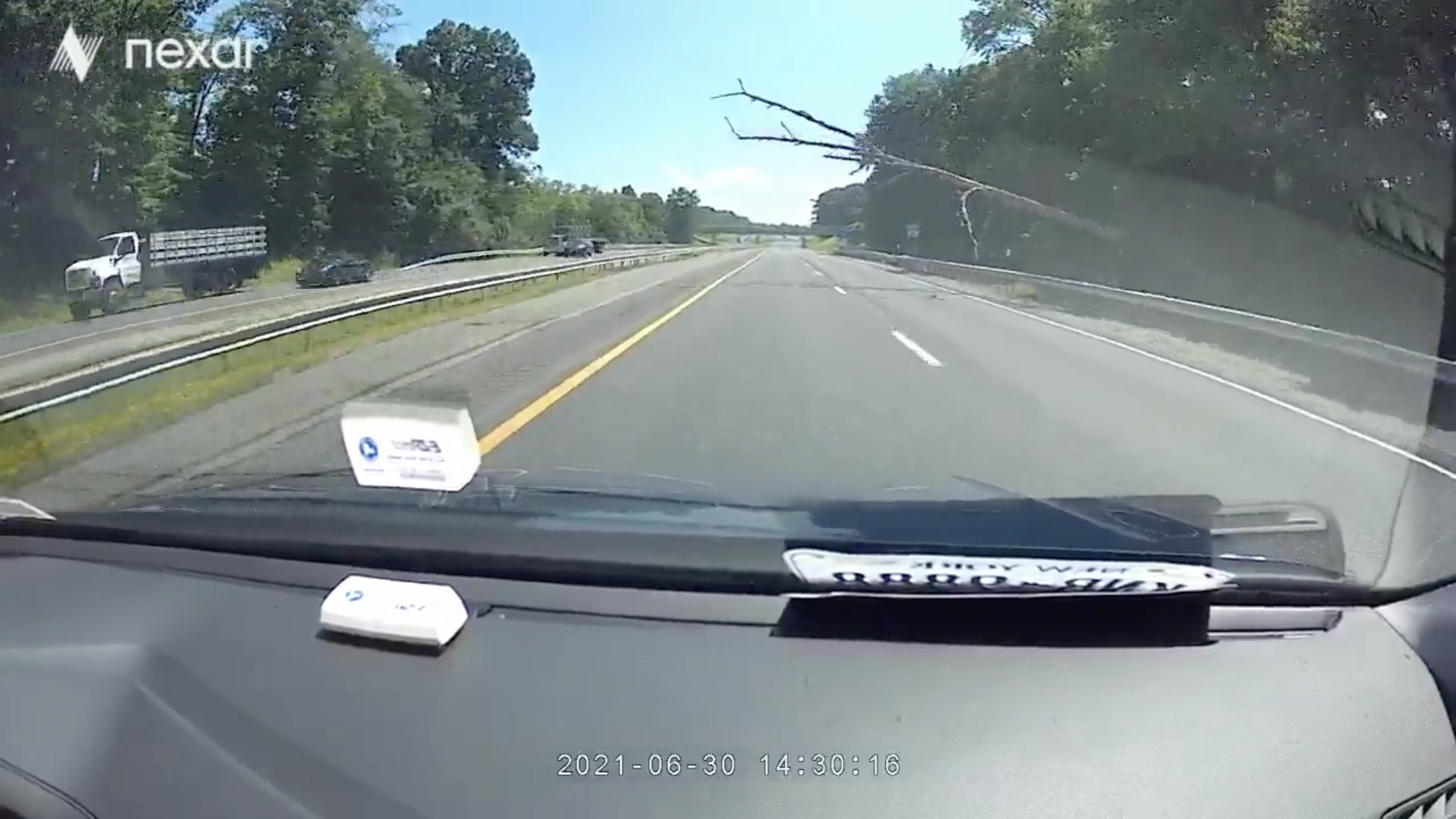} 
    \includegraphics[width=0.24\linewidth]{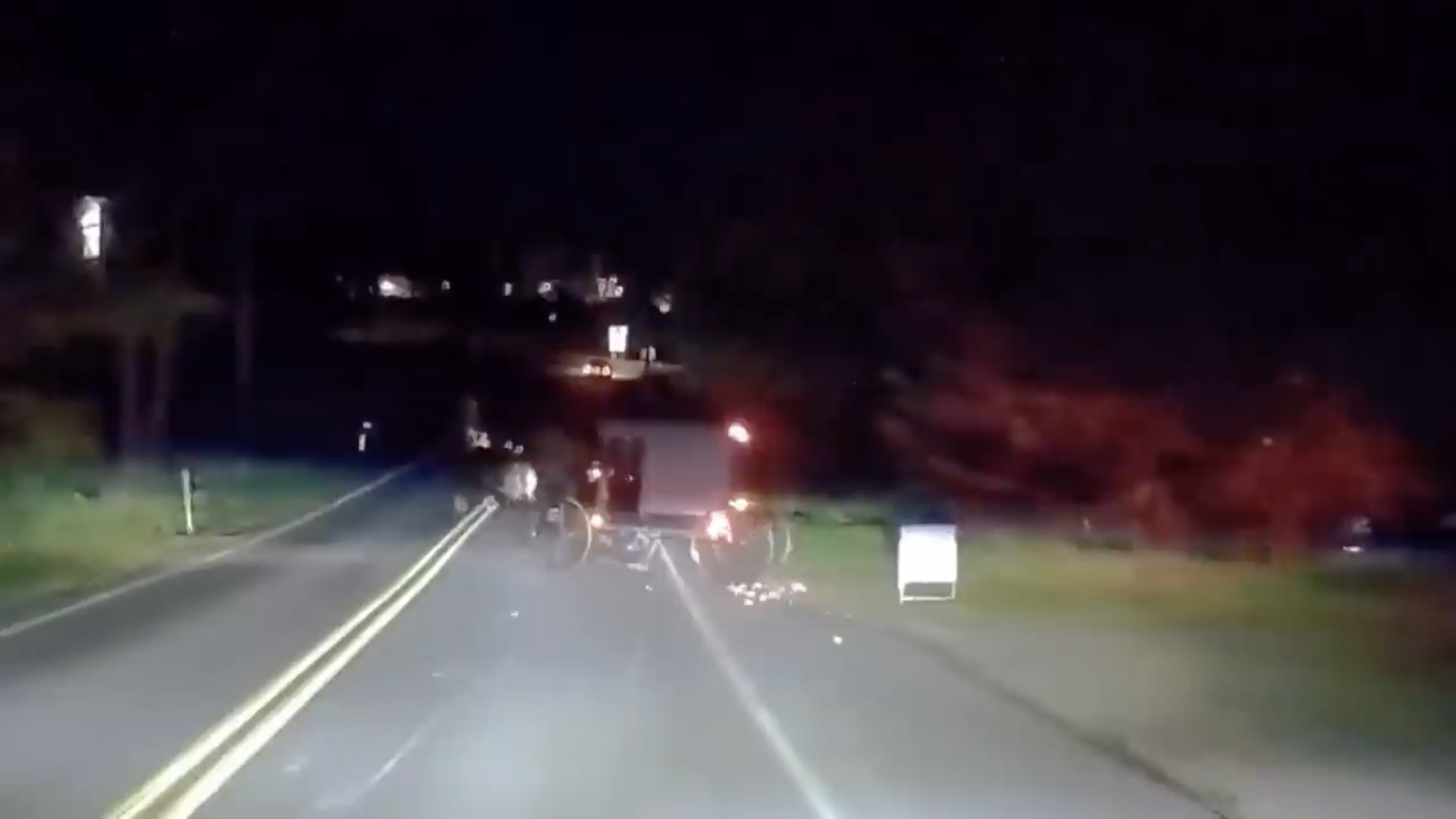} 
    \caption{\textbf{Dashcam views}. The COOOL dataset videos originate from different parts of the world and are of various quality.}
    \label{fig:dataset-samples}
    \vspace{-0.5cm}
\end{figure*}

\begin{figure}[b]
\vspace{-0.25cm}
    \centering
    \includegraphics[height=1.55cm]{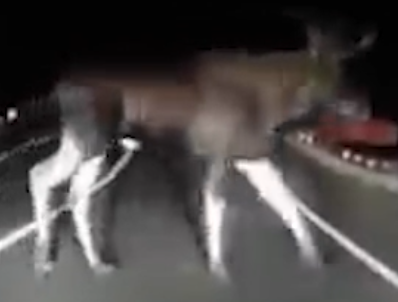}
    \includegraphics[height=1.55cm]{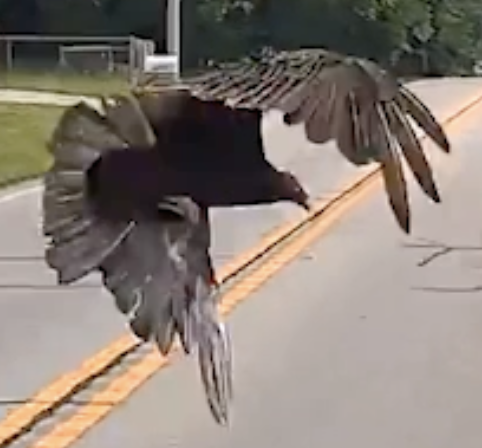}
    \includegraphics[height=1.55cm]{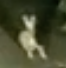}
    \includegraphics[height=1.55cm]{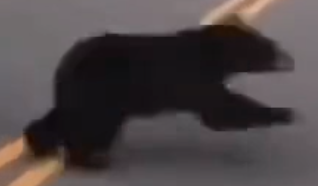} \\
    \vspace{1px}
    \includegraphics[height=1.65cm]{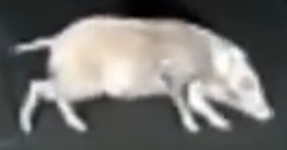}
    \includegraphics[height=1.65cm]{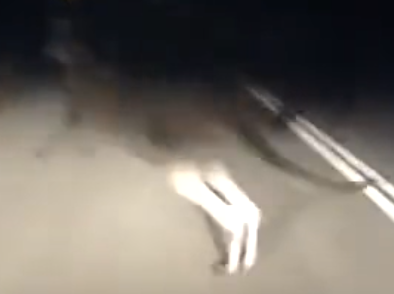}
    \includegraphics[height=1.65cm]{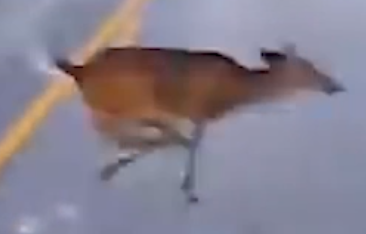} \\
    \vspace{1px}
    \includegraphics[height=1.65cm]{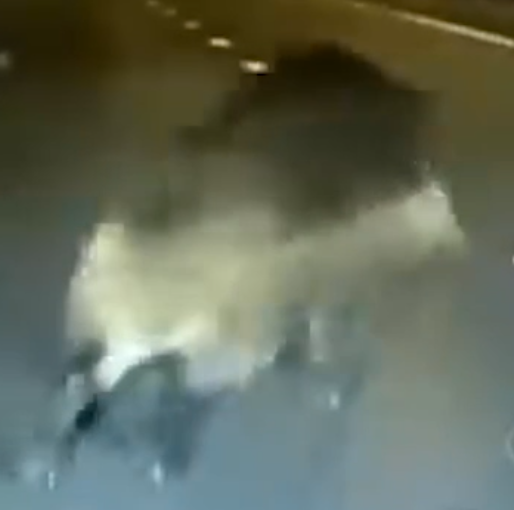} 
    \includegraphics[height=1.65cm]{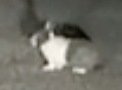}
    \includegraphics[height=1.65cm]{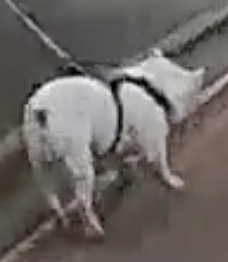}
    \includegraphics[height=1.65cm]{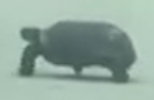} \\
    \vspace{8px}
    \includegraphics[height=1.6cm]{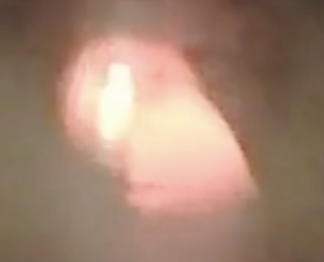} 
    \includegraphics[height=1.6cm]{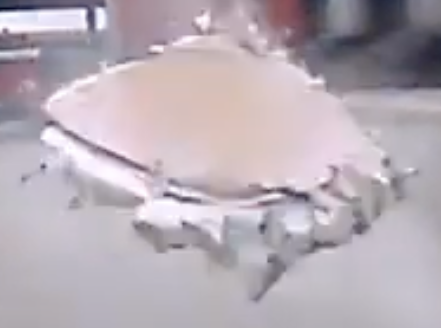}
    \includegraphics[height=1.6cm]{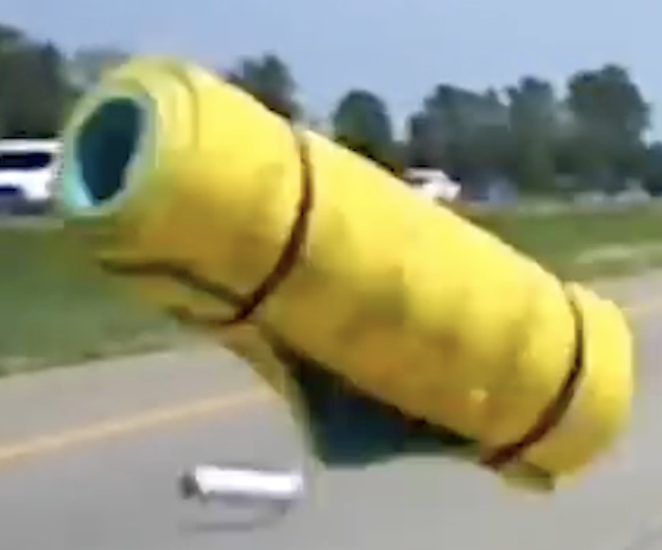} 
    \includegraphics[height=1.6cm]{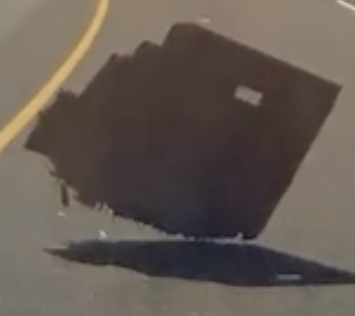} \\
    \vspace{1px}
    \includegraphics[height=1.65cm]{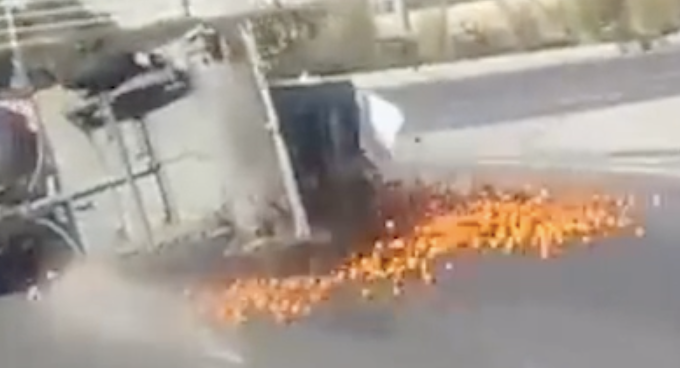}
    \includegraphics[height=1.65cm]{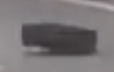}
    \includegraphics[height=1.65cm]{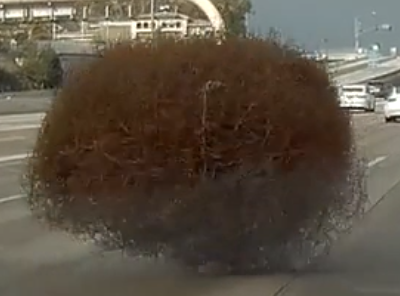} \\
    \caption{\textbf{Cropped out hazard object}. The COOOL dataset includes a wide variety of \textit{animals} (top) and \textit{other} (bottom) objects.}
    \label{fig:hazard-samples}
\end{figure}

\newpage

\subsection{Dataset}
The COOOL dataset is specifically curated to address the detection and classification of out-of-label hazards in autonomous driving. It comprises over 200 dashcam videos of relatively low quality annotated to capture diverse real-world driving scenarios (see Figure \ref{fig:dataset-samples}). The dataset includes a wide spectrum of hazards, categorized into standard hazards — frequently encountered entities such as vehicles, pedestrians, and traffic signs — and uncommon, rare, and hard-to-predict hazards. Uncommon hazards might include exotic animals, such as kangaroos and wild boars, that are region-specific but critical for safety, as well as inanimate objects like plastic bags, fallen tree branches, smoke, and debris that can behave unpredictably or obstruct the road. For reference, see Figure \ref{fig:hazard-samples}. \\

\noindent\textbf{Annotations:} A bounding box and object ID annotations are provided for every frame in each available video. The dataset includes bounding box annotations covering over 100,000 vehicles, 40,000 animals, and various environmental, miscellaneous, and unknown hazards. Each object is assigned a tracklet, enabling the analysis of object trajectories across frames. Most frames feature between 1 and 5 hazards, while some contain up to 26, reflecting the complexity of real-world scenarios.
Among the annotations, over 10,000 are extremely low-resolution objects smaller than 50$\times$50 pixels. These represent distant or very small objects that challenge detection, classification, and captioning methods. 
Additionally, the dataset annotations include driver reaction, registering the timestamp (i.e., frame ID) when drivers responded to hazards (e.g., started braking or steered away).
According to the dataset description, there are between 1 and 18 potential hazard objects, with the majority of videos having three potential hazards present.
Last but not least, the annotations include captions describing the hazards, providing semantic context alongside the visual annotations. These captions specify the nature of the hazard, such as \textit{pedestrian crossing} \textit{vehicle turning}, or \textit{animal on the road}. \\

\noindent\textbf{Note:} \textit{No validation / test split was available for the COOOL competition. None of the annotations were provided for the development. Therefore, we use the public leaderboard as a form of validation set.} \\

\subsection{Evaluation Metric}
The evaluation of the COOOL competition is based on three core tasks: (i) detecting driver reactions, (ii) identifying hazardous objects, and (iii) classifying / captioning hazards. For each task, a score is calculated separately and then combined using an overall macro-accuracy score.

Let the evaluation consist of \( N \) frames in a video sequence, each annotated with ground truth labels. The following metrics are defined:

\noindent\textbf{\textit{Driver reaction accuracy}} (\( A_{reaction} \)): 
assesses the accuracy of detecting the moment the driver reacts to a hazard:

\begin{equation}
A_{reaction} = \frac{1}{N} \sum_{i=1}^{N} \mathbb{I}(r_i = \hat{r}_i),
\end{equation}

where \( r_i \) is the ground truth reaction label for frame \( i \), \( \hat{r}_i \) is the predicted reaction label, and \( \mathbb{I} \) is the indicator function, which equals 1 if \( r_i = \hat{r}_i \), and 0 otherwise. \\

\noindent\textbf{\textit{Hazard identification accuracy}} (\( A_{detection} \)) evaluates the accuracy of identifying objects marked as hazardous:

\begin{equation}
A_{detection} = \frac{1}{N} \sum_{i=1}^{N} \frac{|H_i \cap \hat{H}_i|}{|H_i|},
\end{equation}

where \( H_i \) is the set of ground truth hazardous objects in frame \( i \), \( \hat{H}_i \) is the set of predicted hazardous objects, and \( | \cdot | \) denotes the cardinality of the set. \\

\noindent\textbf{\textit{Hazard classification accuracy}} (\( A_{classific} \))
measures the ability to correctly classify detected hazards:

\begin{equation}
A_{classific} = \frac{1}{N} \sum_{i=1}^{N} \frac{|C_i \cap \hat{C}_i|}{|C_i|},
\end{equation}

where \( C_i \) is the set of ground truth hazard classes in frame \( i \), and \( \hat{C}_i \) is the set of predicted hazard classes. \\

The final evaluation metric is computed as the average of the three individual accuracies:
\begin{equation}
A_{macro} = \frac{\left( A_{reaction} + A_{detection} + A_{classific} \right)}{3}.
\end{equation}

\section{Methods}

This section presents methodologies for driver reaction recognition, hazard identification, and hazard captioning evaluated on the COOOL benchmark. Building on the provided baseline, we introduce an approach based on "old school" computer vision techniques (e.g., optical flow) for motion quantification and bounding box analysis and vision-language models (e.g., BLIP \cite{blip-li2022blip}, CLIP \cite{clip-radford2021learning}, and MOLMO \cite{deitke2024molmo}) for contextual hazard reasoning. \\

\subsection{Baseline}
The baseline method provided by the organizers in the form of a \href{https://github.com/alshami52/COOOL_benchmark}{GitHub repository} represents an interesting approach to the COOOL benchmark. The provided script includes three separate parts, which deal with each metric, i.e., (i) driver reaction recognition, (ii) hazard identification, and (iii) hazard captioning, separately. For clarity, we provide a brief description of the baseline provided below.

\noindent\textbf{Driver reaction recognition:} The provided method calculates the driver's reaction based on \textit{movement patterns} of the annotated objects analyzed throughout the whole video. Divided into steps the method works as follows:
\begin{enumerate}
    \item \textit{Getting objects relative location}: For each object in a video, its location in every frame is determined using the centers from provided bounding boxes.
    \item \textit{Distance calculation}: 
    % Euclidean distances are calculated between the current and previous bounding box centers for all objects and frames. Matching is done using a heuristic that assumes the distance between the centers of the previous and current bounding boxes of the same object is smaller than the distances to other objects. A median of all distance changes is computed, resulting in a single value for each frame.
    For each bounding box center in the current frame, the Euclidean distances to all bounding box centers in the previous frame are calculated. The smallest distance for each bounding box is identified and stored. These minimum distances are then aggregated, and their median is computed to reduce noise and provide a single representative value for the frame. This process is repeated for each frame, resulting in a series of median distances.
    \item \textit{Speed regression}: A linear regression model (using sklearn~\cite{sklearn_api}) is trained to get speed trends. If the trend exhibits a negative slope, it indicates \textit{braking}, which flags a potential change in the driver's state.
\end{enumerate}
\textbf{Note:} \textit{Once driver state change is set, it is set until the end of the video, meaning it is characterized as a step function.} \\

\noindent\textbf{Hazard identification:} The baseline method operates on the hypothesis that hazards are most likely to appear near the center of the screen, representing the area directly in front of the car. To identify the most probable hazard, the bounding box with the center closest to the frame's center is selected. This hazard is associated with a track ID provided in the annotations. While effective for late detections, this approach may struggle with objects that suddenly appear in front of the vehicle or are never there. \\

\noindent\textbf{Hazard captioning:} Once a hazard track is identified, a \textit{chip image} is created by cropping the original image around the bounding box of the identified hazard. A pre-trained visual captioning model is then used to generate a textual description of the cropped image. Captions are generated using the \href{https://github.com/pharmapsychotic/clip-interrogator}{CLIP-Interrogator}, a prompt engineering tool that combines CLIP and BLIP to optimize text prompts to match a given image based on specific phrases or tags from a predefined set called \textit{Flavors} (for more details see ~\cite{clip-inter-udo2023image}). Captions are generated only once when the hazard track is first detected and are reused throughout the rest of the video.

\subsection{Driver reaction recognition}

\begin{figure*}[t]
    \centering
    \includegraphics[width=0.95\linewidth]{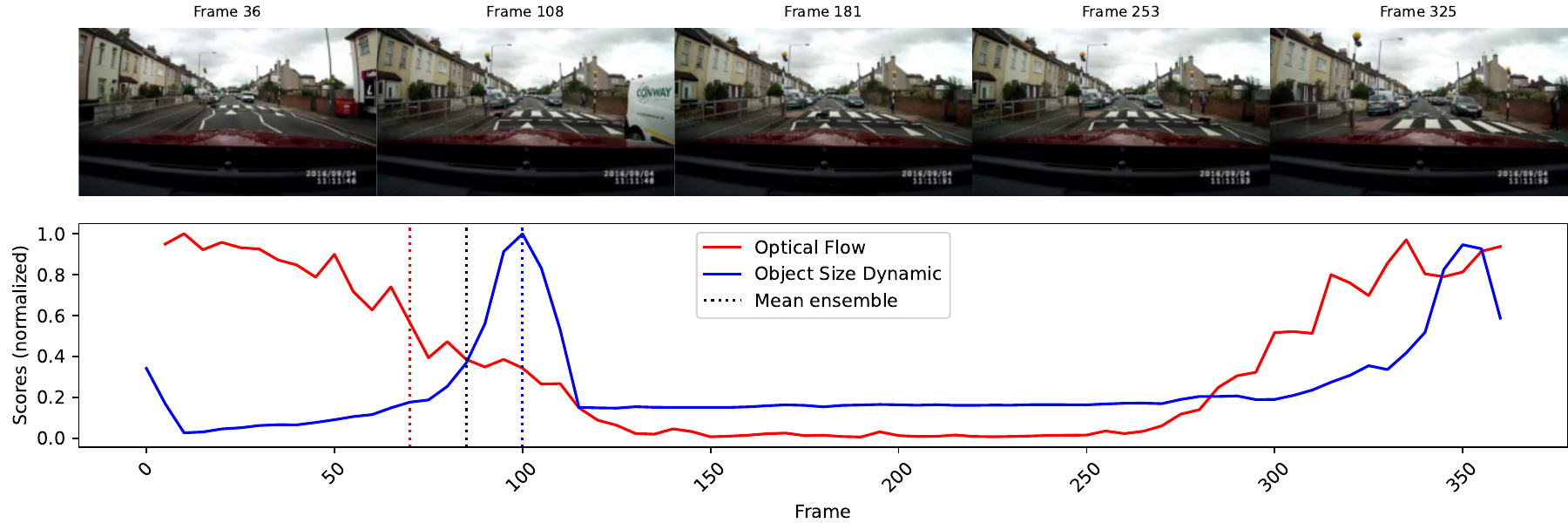}
    \caption{\textbf{Driver reaction recognition}. The optical flow graph demonstrates pixel-level motion intensity over time, with spikes during significant vehicle or background movements. The Object size dynamic graph visualizes the change in bounding box sizes of hazards. Combined (i.e., Mean ensemble in the image), these methods quantify motion and reason about situational risks to identify potential hazards effectively and accurately.}
    \label{fig:driver-change}
\end{figure*}

This section describes methods to identify driver reactions to potential hazards. First, driver state changes are detected using Kernel Change Point Detection applied to the Optical Flow and the Object Size Dynamic. For method overview, see Figure \ref{fig:driver-change}.

\subsubsection{Object size dynamic}
First, we approached the detection of driver state changes by analyzing time series data of total bounding box pixel sizes derived from annotated video frames. This analysis was based on the assumption that a hazardous object becomes larger as it is approached, prompting the driver to slow down. The size of each bounding box, representing an object of interest, was calculated as the pixel area $Bbox\,size = |x_2 - x_1|\cdot|y_2 - y_1|$. The total size per frame was computed by summing the areas of all bounding boxes of the present objects. To account for scale differences across videos, the bounding-box-total-size time series was normalized for each video.

Next, significant changes in the statistical properties of the bounding box size sequence were identified using Kernel Change Point Detection~\cite{celisse2018new, arlot2019kernel} (KernelCPD), implemented with a Radial Basis Function (RBF) kernel using the Ruptures library~\cite{ruptures-truong2020selective}. KernelCPD works by detecting mean shifts in signals mapped to a Reproducing Kernel Hilbert Space (RKHS). To achieve this, the RBF kernel was applied to map the signal into a high-dimensional space, defined as:
\begin{equation}
k(x,y) = e^{-\gamma \cdot {|x-y|}^2},
\end{equation}

where $x,y\,\epsilon\,R^d$ are d-dimensional data points, and $\gamma$ is the bandwidth parameter determined using median heuristics. The detection process involved minimizing a kernelized cost function that measures the dissimilarity of segments to identify breakpoints.

Finally, frames where notable transitions occurred in the bounding box dynamics, suggesting potential driver state changes, were identified through the detected breakpoints. Two solutions are considered: one where the \textit{number of changes} was predefined and another where it was unknown. For scenarios assuming a driver state change in every video, the \textit{number of changes} was treated as an adjustable parameter (set to 4 based on preliminary experiments), with the first breakpoint activating the driver state change prediction. For real-time deployment, however, the mode with an unknown number of breakpoints is more appropriate, as it solves a penalized optimization problem~\cite{killick2012optimal}.

\subsubsection{Optical flow}

Next, we used dense optical flow to estimate inter-frame motion and quantify the vehicle's movement in a video sequence\footnote{Optical flow analyzes pixel intensity changes in consecutive frames to capture motion. Dense optical flow is calculated on a pixel-level.}. The general motivation is that moving vehicles cause many pixels in the frame, especially in the background, to shift. When the vehicle stops, the background remains still, resulting in a drop in the \textit{motion score}. Conversely, when a hazard appears suddenly, there is a rapid change in pixel intensities, reflected as a spike in the motion score. However, optical flow has limitations. It is less reliable in videos with minimal changes in the surrounding environment, such as nighttime footage, where subtle movements are harder to detect. For instance, crossing a bridge at night can create signal artifacts that appear as spikes in the motion score, potentially leading to false hazard detections.

In our experiments, we used the Farneback algorithm~\cite{farneback2003two}, implemented in OpenCV~\cite{itseez2015opencv}. Furthermore, the extracted flow values were converted into polar coordinates to separate the magnitude of motion from its direction. The motion score for each frame was then calculated as the average magnitude of movement across all pixels. By processing the entire video, we obtained a time series that quantitatively represented the vehicle's movement.

% Large bush obstructs road ahead.
% Cat crossing road ahead.
\begin{figure*}[!t]
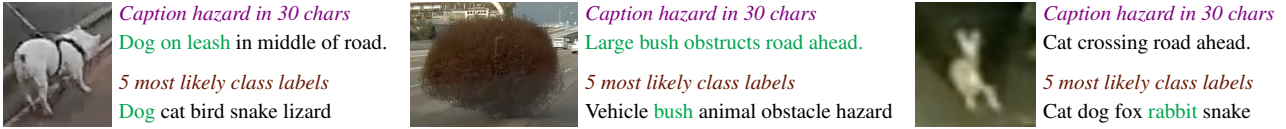

\footnotesize
    \centering
    \begin{tabular}{@{}c@{\hspace{1mm}}l@{\hspace{3mm}}c@{\hspace{1mm}}l@{\hspace{3mm}}c@{\hspace{1mm}}l@{}}
        {\hspace{-1mm}\multirow{4}{*}{\includegraphics[height=1.65cm]{figures/hazards/animals/2.png}}} & \textit{\textcolor{violet}{Caption hazard in 30 chars}} & \multirow{4}{*}{\includegraphics[height=1.65cm]{figures/hazards/other/17.png}} &  \textit{\textcolor{violet}{Caption hazard in 30 chars}} \vspace{0.5mm} & \multirow{4}{*}{\includegraphics[height=1.65cm]{figures/hazards/animals/16.png}} & \textit{\textcolor{violet}{Caption hazard in 30 chars}} \\
        
        &  \textcolor{Green}{Dog on leash} in middle of road.  & & \textcolor{Green}{Large bush obstructs road ahead.} \vspace{2mm} & & Cat crossing road ahead.\\
        & \textcolor{Sepia}{\textit{5 most likely class labels}}  & & \textcolor{Sepia}{\textit{5 most likely class labels}} \vspace{0.5mm} & & \textcolor{Sepia}{\textit{5 most likely class labels}} \\
        &  \textcolor{Green}{Dog} cat bird snake lizard & & Vehicle \textcolor{Green}{bush} animal obstacle hazard & & Cat dog fox \textcolor{Green}{rabbit} snake \\
    \end{tabular}
    \caption{\textbf{Hazard captioning with MOMLO}. Two prompts (1$^{st}$ and 3$^{rd}$ row) and given results. \textit{"Correct"} answer highlighted in green.}
    \label{img:image-captionings}
\end{figure*}

\subsubsection{Ensembling}
To combine the scores, we explore several ensembling schemes. The signals are represented as a time series of Boolean values, starting with \texttt{False} and transitioning to \texttt{True}, after which all subsequent values remain \texttt{True}. The combination methods are described as follows:

\begin{itemize}[left=1mm]
\vspace{-3mm}
    \item \textbf{OR}: A signal is considered \texttt{True} if at least one of the inputs is \texttt{True}. This can be interpreted as a logical union.\vspace{-2mm}
    \item \textbf{AND}: A signal is considered \texttt{True} only if all input signals are \texttt{True}. This can be interpreted as an agreement between the signals.\vspace{-2mm}
    \item \textbf{Mean Position}: For each signal, the position of the first \texttt{True} is identified. The mean of positions across the signals is used as the new position of first \texttt{True}.
\end{itemize}

\subsection{Zero-shot hazard identification}

This section describes methods for identifying hazardous objects without specific training. Two approaches, (i) \textbf{naive} and  (ii) \textbf{object classification}-based, are introduced. The naive approach relies on object proximity to the frame's center, while object classification uses pre-trained models and an area-weighted probability score, excluding predefined safe classes.

\subsubsection{Naive approach} 
The naive approach assumes that all uniquely annotated objects are potentially hazardous according to the annotations' characteristics. Besides, most videos contain at least a few unique objects, and every video is guaranteed to include at least one hazard. Next, any annotated object present in any frame could potentially be classified as hazardous.
Our implementation, thus, relies on this assumption and classifies a given number of tracks as hazardous based on their proximity to the center of the video frame.

\subsubsection{Object classification}
We also approach hazard identification as an object classification problem with a whitelist of classes expected in standard traffic, such as cars or trucks. This strategy considers any object classified outside of the whitelist as a hazard. For the object classification, we use standard out-of-the-shelf models pre-trained on ImageNet \cite{deng2009imagenet} and CIFAR-100 \cite{cifar100}.

For each object in each video frame, we collected the top 10 predicted classes alongside their corresponding softmax probabilities. To determine the final classification of an object across all frames, we applied an \textbf{area-weighted probability score}. This score was calculated as follows:
\begin{equation}
S_c = \sum_{i=1}^{N} P_{c,i} \cdot A_i,
\end{equation}

where $S_c$ is the final score for class $c$, $N$ is the total number of frames in which the object appears, $P_{c,i}$ is the softmax probability of class $c$ for the object in frame $i$ and $A_i$ is bounding box area of the object in frame $i$. The class $c$ with the largest score $S_c$ was selected as the final classification for the object.

\subsection{Hazard captioning}

Following recent advancements in Visual Language Models (VLMs) \cite{blip-li2022blip,achiam2023gpt,alayrac2022flamingo}, we explore their potential for zero-shot hazard identification in autonomous driving, a critical domain requiring a precise understanding of visual inputs. Using the state-of-the-art open-source Malmo-7B VLM model \cite{deitke2024molmo}, we generate hazard annotations by designing prompts tailored to identify and describe potential road hazards focusing on relevance and clarity.

For each object in the video, we extract its five largest images, determined by the area of their bounding boxes, cropped from the original video frames. To avoid introducing additional distortion to the already low-quality, low-resolution images, we ensure that each cropped image is square, with equal height and width. Each object image was captioned using the two prompts:

\begin{itemize}[left=1mm]
\item \textit{"Propose 5 most likely class labels of the object, the context of the image is traffic and unusual hazards such as various animals on the road. Write only the class names separated by spaces."}
\item \textit{"Considering the context of traffic, caption the hazard in one short sentence of max 30 characters and 6 words."
}
\end{itemize}

We generate raw captions for these five images based on the object with the largest bounding box area. From these captions, we extract approximately 25 words in total. The words are then sorted by frequency and, in cases of ties, by their order of appearance in the raw captions. Finally, we concatenate the first five words from this sorted list to create the final caption. For reference, see Figure \ref{img:image-captionings}.

\section{Results}

\subsection{Driver reaction recognition}

The provided baseline approach achieved scores of 0.628 and 0.584 on the public and private sets, respectively, revealing a large room for improvement. In fact, the baseline method scored worse than a naive approach, which assumed that the driver reacted on the first frame of each video (i.e., all values are set to True). On the other hand, the newly proposed Object size dynamic and Optical flow methods performed well, with scores of 0.900 and 0.819, and 0.892 and 0.805, respectively.
By ensembling Optical Flow (OF) and Object Size Dynamic (OSD), the results on the public leaderboard remained more or less the same; however, it slightly improved on the private leaderboard.
For a more comprehensive evaluation, see Table~\ref{tab:metric-1-results}, which summarizes the performance of all evaluated methods for driver reaction recognition.

\begin{table}[h]
\centering
\begin{tabular}{@{}l@{\hspace{2mm}}cc@{}}
\toprule
\textbf{Method}                    & \(A_{reaction}^{public}\) & \(A_{reaction}^{private}\) \\ \midrule
Baseline                           & 0.628 & 0.584       \\
\textit{All True}                  & \textit{0.754} & \textit{0.688}       \\
\textit{All False}                 & \textit{0.246} & \textit{0.310}       \\ \midrule
Object Size Dynamic (OSD)                & 0.900 & 0.819       \\
Optical Flow  (OF)                      & 0.892 & 0.805       \\ \midrule
% MOLMO                              & 0.693 & 0.549       \\ 
(OR)   \,\,\,\,\,OF + OSD                & 0.892 & 0.803       \\
(AND)  OF + OSD                & 0.900 & 0.821       \\
(mean) OF + OSD                & \textbf{0.908} & \textbf{0.829}       \\
% OpticalFlow + Bboxes + MOLMO       & 0.898 & 0.819       \\
\bottomrule
\end{tabular}
\caption{\textbf{Drivers reaction recognition performance.}}
\label{tab:metric-1-results}
\end{table}

\subsection{Hazard identification}

\noindent\textbf{Naive approach:} To evaluate the impact of the number of hazard tracks, we conducted an empirical study with different numbers of nearest tracks. Selecting only the nearest track resulted in a metric score of 0.184 / 0.221 (public / private). Increasing this to the two nearest tracks improved the score to 0.332 / 0.287, while using the three nearest tracks further enhanced the score to 0.405 / 0.311. Finally, including all tracks achieved the highest metric score of 0.799 / 0.535. \\

\noindent\textbf{Object classification:}
In the case of the ImageNet-1k pre-trained classifier, we observed that predicted ImageNet classes were often incorrect, such as pedestrians being classified as "bulletproof vests", likely due to a significant domain shift between the training data and our dataset especially. Additionally, ImageNet consists of a large variety of similar classes, making it hard to create a whitelist of non-hazard classes. For the most frequently classified categories, see Table \ref{tab:classifications}.

Instead, we opted to use the same ViT-Base model \cite{wu2020visual} (with patch size 16 and 224$\times$224 input resolution), which was pre-trained on ImageNet and fine-tuned on CIFAR-100 \cite{cifar100}. This approach yielded much more reasonable classifications. For example, pedestrians were accurately classified as $man$, $woman$, $girl$, or $boy$.  We define a hazard as any class that is not one of the following: $pickup truck$, $bus$, $tank$, $motorcycle$, or $cloud$.  We included the $tank$ and $cloud$ classes because they are frequently used to represent vehicles with small bounding boxes, particularly in nighttime or low-visibility conditions.

Filtering out only cars results in a private leaderboard score of 0.474, representing a significant improvement (+0.119) compared to using all tracks. Excluding certain classes from the whitelist further increases the score to 0.535. Additionally, applying a simple trajectory size filter—removing objects with a trajectory smaller than their width or height—yields another notable improvement (+0.035) on the private leaderboard.
By combining both the whitelist and the size filter, we achieve our best private leaderboard score of 0.570. For a more comprehensive evaluation of the achieved results, see Table \ref{tab:metric-2-results}.

\begin{table}[h]
\centering
\begin{tabular}{ll|ll}
\toprule
\multicolumn{2}{c|}{\textbf{CIFAR-100}} & \multicolumn{2}{c}{\textbf{ImageNet-1k}} \\
Category & Freq. (\%) & Category & Freq. (\%) \\
\midrule
pickup & 20.1 & traffic light & 16.4 \\
bus & 17.2 & trailer truck & 13.7 \\
cloud & 5.8 & car mirror & 12.6 \\
man & 5.4 & lakeside & 4.5 \\
tank & 3.4 & street sign & 4.3 \\
road & 2.6 & pole & 4.2 \\
lamp & 2.6 & goose & 3.6 \\
motorcycle & 2.6 & garbage truck & 3.4 \\
shark & 2.0 & moving van & 3.2 \\
seal & 1.9 & pickup & 3.1 \\
\bottomrule
\end{tabular}
\caption{\textbf{Top 10 most frequently predicted categories.} Compared to the CIFAR-100 fine-tuned model, the ImageNet-1k model predictions are often way off from the correct category.}
\label{tab:classifications}
\vspace{-0.25cm}
\end{table}
\begin{table}[h]
\centering
\begin{tabular}{@{}lcc@{}}
\toprule
\textbf{Method}                    & \(A_{detection}^{public}\) & \(A_{detection}^{private}\) \\ \midrule
1 Nearest Track  -- \textit{Baseline}         & 0.184 &  0.221\\ %results_10122024_baseline_driver_track.csv - results_10122024_baseline_driver.csv
2 Nearest Tracks                    & 0.332 &  0.287 \\ % results_11122024_bboxsizes_driver_2tracks.csv - results_09122024_bboxsizes.csv

3 Nearest Tracks                    & 0.405 & 0.311  \\ %results_11122024_bboxsizes_driver_3tracks.csv - results_09122024_bboxsizes.csv
All Tracks                          & 0.618 & 0.355  \\ %results_11122024_bboxsizes_driver_alltracks.csv - results_09122024_bboxsizes.csv
\midrule
All Tracks without cars       & 0.686 & 0.474 \\ % results_11122024_bboxsizes_driver_alltracks_molmo-cls-filter.csv - results_09122024_bboxsizes.csv
All + Object classification filter  & \textbf{0.799} & \textbf{0.535}        \\ %results_11122024_driver_true_alltracks_cifar-cls-filter.csv - baseline_fixes.csv
\textit{Above} + Size filter  & \textbf{0.799} & \textbf{0.570}       \\%results_18122024_driver_best_alltracks_cifar-filter-molmo-caps-filtersize.csv - 
% All + Cifar filter + Car Filter    & 0.25355                  & 76.065       \\ \hline
All + MOLMO filter                  & 0.761 &  0.568  \\ % results_17122024_driver_true_alltracks_cifarmolmo-cls-filter.csv - baseline_fixes.csv
\bottomrule
\end{tabular}
\caption{\textbf{Hazard identification performance.}}
\label{tab:metric-2-results}
\end{table}

\subsection{Hazard captioning}
Table~\ref{tab:metric-3-results} show that the MOLMO-based methods significantly outperform the provided baseline in hazard captioning and its extended version with All Tracks. The "1 Nearest Track" and "All Tracks" baselines achieve public scores of 0.022 and 0.065, respectively, indicating limited performance.  
The "MOLMO -- \textit{Sentence}" approach (\(A_{classific}^{public}=0.062\)) shows no substantial improvement over the baselines, suggesting sentence-based prompts may not capture hazard context effectively. In contrast, the "MOLMO -- \textit{Categories}" method achieves the best results (\(A_{classific}^{public}=0.239\), \(A_{classific}^{private}=0.162\)), highlighting the efficacy of category-based prompts tailored for traffic.

\subsection{Competition evaluation}
For the final competition evaluation, we selected two runs\footnote{Only two runs were permitted for the final evaluation on Kaggle.}. The first one was based on the best-performing approaches for each task.  
For driver reaction recognition, we utilized an ensemble combining Optical Flow with Object Size Dynamics (\(A_{reaction}^{private} = 0.821\)).  
For hazard identification, we selected all objects and filtered out those classified as not fitting standard traffic object categories (\(A_{detection}^{private} = 0.535\)).
For hazard captioning, we employed a MOLMO model prompted to predict the five most likely CIFAR-like categories (\(A_{classification}^{private} = 0.162\)).

The second run included the removal of objects with trajectories shorter than their bounding box width or height, indicating extremely limited movement (\(A_{detection}^{private} = 0.570\)). 
The second run achieved a slightly bigger score (0.51772) in terms of the official metric and secured 2\textsuperscript{nd} place in the competition. For a final standing, please refer to Table \ref{tab:leaderboard}.

\section{Conclusion}
In this work, we presented our approach to the COOOL competition, addressing the detection and classification of out-of-label hazards in autonomous driving. Our pipeline combined traditional computer vision techniques with cutting-edge vision-language models to work relatively well in all the tasks, i.e., in driver reaction detection, hazard identification, and hazard captioning. The integration of kernel-based change point detection, optical flow, and pre-trained vision and vision language models proved highly useful, culminating in a 33\% reduction in relative error compared to baseline methods and securing 2nd place among 32 competing teams.

Despite the good performance, our approach has notable \textbf{limitations}. 
First, while partially mitigated by our methods, the low-quality / low-resolution inputs still impact performance, particularly in hazard captioning. 
Second, our reliance on pre-trained models introduces domain shift challenges, which may limit generalizability to datasets with significantly different characteristics. 
Finally, the lack of ground truth validation during the competition constrained model tuning, potentially affecting optimal performance.

\textbf{Future work} will focus on enhancing robustness to domain variations, leveraging self-supervised techniques to improve generalization, and exploring real-time inference to address practical deployment challenges further.

\begin{table}[h]
\centering
\begin{tabular}{@{}lcc@{}}
\toprule
\textbf{Method}                    & \(A_{classific}^{public}\) & \(A_{classific}^{private}\) \\ \midrule
1 Nearest Track  -- \textit{Baseline}        & 0.022 & 0.009    \\ % results_10122024_baseline_driver_track_cap.csv - results_10122024_baseline_driver_track.csv
All Tracks    & 0.065 &  0.016 \\ % results_13122024_bboxsizes_driver_alltracks_allcaps.csv - results_11122024_bboxsizes_driver_alltracks.csv
MOLMO -- \textit{Sentence}           & 0.062 & 0.069 \\ % results_11122024_bboxsizes_driver_alltracks_molmo-cls-filter-molmo-capshort.csv - results_11122024_bboxsizes_driver_alltracks_molmo-cls-filter.csv
% 0.54751, 0.45040 - 0.52680, 0.42733
MOLMO -- \textit{Categories}         & \textbf{0.239} & \textbf{0.162}   \\ \bottomrule % results_18122024_driver_best_alltracks_cifar-filter-molmo-caps.csv - 
\end{tabular}
\caption{\textbf{Hazard captioning performance.}}

\label{tab:metric-3-results}
\end{table}

\begin{table}[h]
\centering
\begin{tabular}{clcc}
\toprule
\# & \textbf{Team name}                   & \(A_{reaction}^{public}\) &\(A_{reaction}^{private}\) \\ \midrule
1 &Duong Anh Kiet                      & 0.78453 & 0.57261 \\
2 &\textit{(our)}                      & 0.63993 & 0.51772 \\
3 &Impish                              & 0.68794 & 0.51596 \\
4 &Parisa Hatami                       & 0.54599 & 0.48967 \\
5 &TeamCV                              & 0.55705 & 0.44401 \\
6 &PMM\_UTCU                            & 0.43161 & 0.44020 \\
7 &Mahdi Abbariki                      & 0.56956 & 0.37568 \\
8 &Nachiket Kamod                      & 0.43368 & 0.31733 \\
9 &Peace.LU                            & 0.34695 & 0.31639 \\
10 &john                                & 0.31198 & 0.29685 \\
\bottomrule
\end{tabular}
\caption{\textbf{COOOL challenge -- Final standing -- Top10 teams.}}
\label{tab:leaderboard}
\end{table}

%%%%%%%%% REFERENCES
{\small
\bibliographystyle{ieee_fullname}
\bibliography{references}
}

\end{document}